\documentclass[10pt,twocolumn,letterpaper]{article}

\usepackage{iccv}
\usepackage{times}
\usepackage{epsfig}
\usepackage{graphicx}
\usepackage{amsmath}
\usepackage{amssymb}

\usepackage{subfig}
\usepackage{multirow}
\usepackage{array}
\usepackage{multirow}

\usepackage[pagebackref=true,breaklinks=true,letterpaper=true,colorlinks,bookmarks=false]{hyperref}

\iccvfinalcopy 

\def\iccvPaperID{3233} 

\begin{document}

\title{Bilateral Adversarial Training: Towards Fast Training of More Robust Models Against Adversarial Attacks}

\author{Jianyu Wang \\
Baidu Research USA \\
{\tt\small wjyouch@gmail.com}
\and
Haichao Zhang \\
Baidu Research USA \\
{\tt\small hczhang1@gmail.com}
}

\maketitle
\ificcvfinal\thispagestyle{empty}\fi

\begin{abstract}
In this paper, we study fast training of adversarially robust models. From the analyses of the state-of-the-art defense method, i.e., the multi-step adversarial training~\cite{madry2017towards}, we hypothesize that the gradient magnitude links to the model robustness. Motivated by this, we propose to perturb both the image and the label during training, which we call Bilateral Adversarial Training (BAT). To generate the adversarial label, we derive an closed-form heuristic solution. To generate the adversarial image, we use one-step targeted attack with the target label being the most confusing class. In the experiment, we first show that random start and the most confusing target attack effectively prevent the label leaking and gradient masking problem. Then coupled with the adversarial label part, our model significantly improves the state-of-the-art results. For example, against PGD100 white-box attack with cross-entropy loss, on CIFAR10, we achieve 63.7\% versus 47.2\%; on SVHN, we achieve 59.1\% versus 42.1\%. At last, the experiment on the very (computationally) challenging ImageNet dataset further demonstrates the effectiveness of our fast method.

\end{abstract}

\section{Introduction}
Deep learning has achieved great success in many visual recognition tasks in computer vision. However, deep neural networks are extremely vulnerable to adversarial attacks ~\cite{szegedy2013intriguing}. Specifically, the network is easily fooled to make wrong predictions in the face of adversarial examples, which are adversarially manipulated images by adding small and imperceptible perturbations. This poses a great danger to deploying real-world machine learning systems. Therefore, training an adversarially robust model is of great value towards commercialized AI technology.

In the recent years, many approaches have been proposed to defend against adversarial examples. As demonstrated by~\cite{athalye2018obfuscated}, multi-step adversarial training~\cite{madry2017towards} is currently the best defense method. Specifically, adversarial training solves a minimax (saddle point) problem. The inner maximization generates adversarial examples by multi-step projected gradient descent (PGD), which are then used in the outer minimization to optimize the network parameters.

To understand the working mechanism of the multi-step adversarial training, we first perform two diagnostic experiments on CIFAR10. The first experiment is to test a seemingly correct assumption: using stronger adversarial attacks during training will lead to more robust models. To this end, we compare two adversarially trained models, which only differ in the hyper-parameters of the inner maximization: 1) the default setting in~\cite{madry2017towards}, denoted by PGD7-2, where the number of iterations is 7 and the step size is 2 pixels; 2) a model trained using a weaker attack, denoted by PGD2-8, where the number of iteration is 2 and the step size is 8 pixels. 
We observe that PGD2-8 is largely as robust as PGD7-2 under different white-box attacks, even though PGD2-8 attack is weaker than PGD7-2 attack. This result leads us to hypothesize that robustness may not be achieved by simply fitting sufficient adversarial examples during training, and to re-consider if there is more essential ingredients that directly relate to network robustness. With this in mind, we conduct the second experiment where we compare the gradient magnitude of both undefended models and adversarially trained models. We observe that the gradient magnitude of adversarially trained models is much smaller than that of the undefended models. Intuitively speaking, if the gradient (with respect to input images) becomes extremely small, the gradient-based adversarial attacks are likely to fail no matter how many iterations are used. This inspires us that gradient magnitude may directly link to model robustness.

Based on the above observations, in order to achieve adversarial robustness, we would like a model to satisfy the following two conditions: 1) low loss (zero-order condition); 2) small gradient magnitude (first-order condition). To this end, in this paper, we propose a formulation to achieve these two conditions by perturbing both input images and labels during training, which we call Bilateral Adversarial Training (BAT). As for generating the adversarial image, we adopt one-step PGD which speeds up the training by multiple times compared to~\cite{madry2017towards}. In order to avoid the troublesome label leaking and gradient masking problem often induced by one-step PGD~\cite{kurakin2016scale,tramer2017ensemble}, we do the following: 1) using targeted attack with the target label being the most confusing class; 2) adding random uniform noise to the original image as initialization, i.e., random start as in~\cite{madry2017towards}. As for generating the adversarial label, we derive a formula to perturb the groundtruth label (in the form of one-hot vector) based on the gradient with respect to input label (i.e., the negative logarithm probability). As a special case, this solution reduces to label smoothing when the gradient of non-groundtruth classes are equal. 

In the experiment, we first empirically demonstrate that random start and the MC targeted attack are very effective at avoiding label leaking and gradient masking problem. The model trained by using these two techniques alone can achieve similar robustness as multi-step adversarially trained models in~\cite{madry2017towards}. Next, after adding adversarial label part, our model significantly improves the state-of-the-art results in~\cite{madry2017towards}. In order for rigorous robustness evaluation, we use the strong white-box attacks such as PGD100 and PGD1000 with both cross-entropy loss and margin-based loss. For example, against PGD100 under cross-entropy loss, on CIFAR10, we achieve 63.7\% versus 47.2\%; on SVHN, we achieve 59.1\% versus 42.1\%. At last, we apply our fast method to the very challenging ImageNet dataset. Our model is successfully trained using only 8 GPUs, compared with 53 GPUs~\cite{kannan2018adversarial} and 128 GPUs~\cite{xie2018feature}. Compared with the recent state-of-the-art~\cite{xie2018feature}, our model is better on clean images and against non-targeted attacks, but worse against randomly targeted attacks, using an order-of-magnitude less computational resources.

In summary, our contribution is threefold. First, we empirically show that small gradient magnitude may improve the adversarial robustness. Second, we propose a fast adversarial training method called BAT, which perturbs both the image and the label. Third, our method significantly improves the state-of-the-art results on several datasets. 


\section{Related Work}
\subsection{Adversarial Attacks}
Adversarial examples have long been studied in machine learning~\cite{dalvi2004adversarial,huang2011adversarial,biggio2013evasion,biggio2018wild}. In the time of modern deep learning, \cite{szegedy2013intriguing} first pointed out that CNNs are vulnerable to adversarial examples, and proposed a box-constrained L-BFGS method to compute them. Later on, \cite{goodfellow2014explaining} proposed the fast gradient sign method (FGSM) to efficiently generate adversarial examples. FGSM was then extended to an iterative version in~\cite{kurakin2016adversarial}, which showed that adversarial examples can exist in the physical world. In~\cite{moosavi2015deepfool}, the authors proposed DeepFool to compute the adversarial perturbations, and define and quantify the robustness of classifiers. In~\cite{carlini2016towards}, the famous CW attack was proposed, which used the margin-based loss, and applied change-of-variables to remove the constraint. In spite of being very slow, CW attack is currently one of the strongest attacks. Later \cite{chen2017ead} modified the loss function in~\cite{carlini2016towards} by applying elastic net regularization. 

There are some works devoted to improving the transferability of adversarial examples, which leads to stronger black-box attacks. \cite{liu2016delving} proposed to compute the adversarial perturbation by attacking an ensemble of networks simultaneously, and demonstrated improved transferability. In~\cite{papernot2017practical}, the authors assumed a scenario where the attackers have access to the prediction results of a few examples. They then trained a substitute/surrogate model based on the limited number of examples, and generate adversarial examples using the substitute model. \cite{dong2017boosting} demonstrated that momentum-based iterative attacks achieve better transferability. There are some works proposing zeroth-order attacks, i.e., using the logit to generate the attacks~\cite{uesato2018adversarial,chen2017zoo}. Besides, \cite{brendel2017decision} proposed the boundary attack, which is based on the final model decision, instead of the gradient or logit.

In addition to image classification, adversarial examples were also studied in many other tasks, including object detection~\cite{xie2017adversarial}, semantic segmentation~\cite{xie2017adversarial,metzen2017universal}, speech recognition~\cite{cisse2017houdini}, image captioning~\cite{chen2017show}, deep reinforcement learning~\cite{huang2017adversarial,pinto2017robust}. Besides the additive perturbation model, \cite{engstrom2017rotation} studied how to generate adversarial examples under rotation and translation. \cite{evtimov2017robust} studied physically adversarial example in the context of detecting stop sign in real world. Another interesting topic is given by~\cite{athalye2017synthesizing}, where the authors synthesized robust adversarial examples in 3D. 

\subsection{Adversarial Defenses}
In recent years, many methods have been proposed to defend against adversarial examples. One line of research is on detecting adversarial examples, such as~\cite{metzen2017detecting,meng2017magnet}. But later \cite{carlini2017adversarial} showed that their CW attack is able to bypass most detection methods. Another line of research tries to break the special structure in adversarial perturbation by random or non-differentiable operations \cite{xie2017mitigating,guo2017countering,song2017pixeldefend,samangouei2018defensegan,liao2017defense,prakash2018deflecting,liu2017towards}. Recently, \cite{athalye2018obfuscated} showed that many existing defense methods relied on gradient masking, which leads to a false sense of robustness against adversarial attacks. Besides, gradient-based regularization~\cite{jakubovitz2018improving,ross2017improving} and nearest neighbor~\cite{dubey2019defense} have been demonstrated to improve robustness.

Adversarial training~\cite{goodfellow2014explaining,kurakin2016scale,tramer2017ensemble,madry2017towards,sinha2018certifying,raghunathan2018certified,xie2018feature,zhang2019theoretically} is currently the best defense method against adversarial attacks. \cite{kurakin2016scale} first scaled up adversarial training to ImageNet dataset, where the authors used one-step least likely targeted attack to generate adversarial examples during training. Later in~\cite{tramer2017ensemble}, the authors pointed out that such adversarially trained models suffer from gradient masking, and proposed ensemble adversarial training, which augmented the training data with perturbations computed from a set of held-out models. \cite{madry2017towards} demonstrated that multi-step adversarial training is very effective at achieving robustness, and also managed to avoid the gradient masking problem. According to~\cite{athalye2018obfuscated}, this is currently the best defense method.

\section{Motivation}
In this section, we empirically analyze two aspects of the multi-step adversarial training method in~\cite{madry2017towards}: 1) if more iterations in the inner maximization improves the robustness, and 2) the gradient magnitude of both undefended models and adversarially trained models. The experiments are conducted on CIFAR10. Based on the analyses, we hypothesize that making the loss surface locally flat (i.e., small local gradient magnitude) helps achieve better robustness. The proposed algorithm will then be given in the next section.

\subsection{Background}
We first give a short description of the adversarial training method in~\cite{madry2017towards}. This method achieves the currently best adversarial robustness according to~\cite{athalye2018obfuscated}. Specifically, it solves the following saddle point (minimax) problem
\begin{equation}
\min_{\theta} \{\mathbb{E}_{(x,y) \sim \mathcal{D}}[\max_{x' \in \mathcal{S}_x} L(x', y;\theta)]\}.
\end{equation}
Here $(x, y)$ denotes the original data point, $x'$ denotes the adversarially perturbed image, $L(\cdot)$ denotes the loss function, and $\epsilon_x$ denotes the perturbation budget. The feasible region $\mathcal{S}_x$ is defined as~\footnote{In implementation, we rescale all images with pixel values in [-1, 1].}
\begin{equation}
\mathcal{S}_x=\{z\,|\, \ z \in B(x, \epsilon_x) \cap [-1.0, 1.0]^n\},
\label{s_x}
\end{equation}
where $B(x, \epsilon_x)\!\triangleq\!\{z\,|\,\|z - x\|_{\infty} \leq \epsilon_x\}$ denotes the \mbox{$\ell_{\infty}$-ball} with center $x$ and radius $\epsilon_x$.
In the following, for the sake of notational simplicity, without loss of generality, we present the formulation based on a single sample.
The outer minimization is minimizing the cross-entropy loss as in the standard classification. The inner maximization corresponds to the adversarial attack. In order to better explore the solution in $B(x, \epsilon_x)$, \cite{madry2017towards} uses random start before taking a number of PGD steps, i.e., 
\begin{align}
x^0 &\sim B(x, \epsilon_x), \label{random}\\
x^{t+1} &= \Pi_{\mathcal{S}_x}\big(x^t + \epsilon_x \cdot \text{sign}\big(\nabla_{x}L(x^t, y; \theta)\big)\big). \label{PGD}
\end{align}
The original image $x$ is at first randomly (uniform) perturbed to some point $x^0$ in $B(x, \epsilon_x)$ as in (\ref{random}), and then goes through several PGD steps as in (\ref{PGD}). The $\Pi_{\mathcal{S}_x}(\cdot)$ operator projects the input into the feasible region $\mathcal{S}_x$.

\begin{table}[t!]
\centering
\tabcolsep=0.055cm
\begin{tabular}{ |c||c|c|c|c|c|c| }
\hline
Acc.(\%) & clean & \small FGSM & \small PGD2-8 & \small PGD7-2 & \small PGD20-2 & \small PGD100-2 \\
\hline
\small PGD7-2   & 88.0	& 57.0 & 53.0 & 51.2 & 47.6 & 47.2  \\
\hline
\small PGD2-8   & 88.2	& 56.9 & 53.2 & 50.5 & 46.7 & 46.2  \\
\hline
\end{tabular}
\caption{Comparison between the model in~\cite{madry2017towards} (top) and an adversarially trained model using a weaker attack during training (bottom). They achieve similar robustness.}
\label{motive1}
\end{table}

\begin{table}[t!]
\centering
\tabcolsep=0.05cm
\begin{tabular}{ |c||c|c|c||c|c|c| }
\hline
\multirow{2}{*}{} & \multicolumn{3}{c||}{undefended}
                  & \multicolumn{3}{c|}{adversarially trained} \\
\cline{2-7}
                 & \multicolumn{1}{c|}{all} 
                 & \multicolumn{1}{c|}{correct}
                 & \multicolumn{1}{c||}{wrong}
                 & \multicolumn{1}{c|}{all}
                 & \multicolumn{1}{c|}{correct}
                 & \multicolumn{1}{c|}{wrong} \\
\hline
min     & 3.0e-32  & 3.0e-32  & 264.1   & 2.6e-26  & 2.6e-26  &  0.2    \\
\hline
mean    & 395.0    & 23.6     & 7.4e3   & 3.8      & 0.4      &  28.9   \\
\hline
max     & 4.5e4    & 7.0e3    & 4.5e4   & 236.2    & 85.9     &  236.2  \\
\hline
\end{tabular}
\caption{The minimal, average, and maximal value of gradient magnitude of the test images on CIFAR10. Overall Adversarially trained models have much smaller gradient magnitude than undefended models. All models are trained using the same regularization, epoch, learning rate, etc.}
\label{motive2}
\vspace{-0.3cm}
\end{table}

\subsection{Analyses}
{\flushleft\textbf{Do more iterations help?}}
We first examine if more iterations in inner maximization help improve the robustness. To this end, we compare two adversarially trained models with different hyper-parameters for generating the adversarial examples during training. The first one is the default in~\cite{madry2017towards}, denoted by PGD7-2, which uses 7 steps of PGD, and step size is 2.0. The second one is a seemingly weaker variant, denoted by PGD2-8, meaning only 2 steps of PGD are used and step size is 8.0. As in~\cite{madry2017towards}, the perturbation budget is 8.0 in training and evaluation, and random start is used. From Table~\ref{motive1}, we see that PGD2-8 performs roughly the same as PGD7-2, against PGD attacks with different steps (strength). This result leads us to hypothesize that using stronger attacks during training may not necessarily lead to more robust models.
{\flushleft\textbf{Gradient magnitude of adversarially trained models.}}
Next we examine the gradient magnitude of the undefended models and adversarially trained models. We consider three collections of all test images in CIFAR10, and for each collection we compute the minimal, average, and maximal of gradient magnitude, i.e., $\| \nabla_{x}L(x, y; \theta) \|^2_2$. The three collections are: 1)~entire images, denoted by ``all", 2)~correctly predicted images, denoted by ``correct", 3)~wrongly predicted images, denoted by ``wrong". The numbers are shown in Table~\ref{motive2}. First we see that for any collection, the gradient magnitude of undefended model is much larger than that of adversarially trained model. Also, for each model, the gradient magnitude of correctly predicted images is much smaller than that of wrongly predicted images.

\subsection{Hypothesis}
From the above analyses, we hypothesize that small gradient magnitude directly links to the adversarial robustness. Intuitively speaking, if the loss surface is locally ``flat" around the data points, the model is hard to attack, no matter how many steps are used. This hypothesis is aligned with~\cite{liu2017towards} where they call it local Lipschitz. Note that there are papers~\cite{hochreiter1997flat,dinh2017sharp} studying the possible relationship between the flatness of loss surface and the generalization of the model. In this paper, we simply use gradient magnitude to measure the ``flatness". A rigorous treatment is beyond the scope of the current paper and left as future work.

A straightforward idea to reduce the gradient magnitude is to augment the loss function with some form of gradient regularization during training, e.g., \cite{ross2017improving,jakubovitz2018improving}. However, the key problem of this idea is that training requires the computation of second-order derivatives, which becomes extremely slow and expensive for large-scale networks.

\section{Formulation}
In this section, in order to improve the adversarial robustness, we propose Bilateral Adversarial Training (BAT), which simultaneously perturbs both the image and the label during adversarial training.

We first approximately relate the general adversarial training framework to small gradient magnitude. Let $x, x'$ denote the original and the slightly perturbed image, and $y, y'$ denote the original and the slightly perturbed groundtruth (in the form of a probability distribution lying in the probability simplex). Let $L(\cdot)$ denote the loss function. The first-order Taylor expansion of the loss is
\begin{equation}
\begin{split}
L(x', y'; \theta) \, \approx \, & L(x, y; \theta) \\
& + \nabla_xL(x,y;\theta) \cdot (x' - x) \\
& + \nabla_yL(x,y;\theta) \cdot (y' - y).
\end{split}
\label{taylor}
\end{equation}
We use the perturbation budget constraint in $\mathcal{\ell}_{\infty}$-norm, i.e., 
\begin{equation}
\| x' - x \|_{\infty} \leq \epsilon_x, \; \| y' - y \|_{\infty} \leq \epsilon_y.
\end{equation}
By Holder's inequality, from (\ref{taylor}) we can approximately have the upper bound 
\begin{equation}
\begin{split}
L(x', y';\theta) \, \leq \, & L(x,y;\theta) \\
&  +  \epsilon_x \|\nabla_xL(x,y;\theta)\|_1\\
& + \epsilon_y \|\nabla_yL(x,y;\theta)\|_1.
\end{split}
\label{holder}
\end{equation}


Intuitively speaking, adversarial training, by minimizing $L(x', y'; \theta)$, translates to 1) minimizing $L(x, y; \theta)$ and 2) minimizing the gradient magnitude $\| \nabla_xL(x,y;\theta) \|_1$ and $\| \nabla_yL(x,y;\theta) \|_1$. The second point explains the results in Table~\ref{motive2}. Note that the first point makes the network predict the correct class, and the second point makes it difficult to generate adversarial examples for gradient-based attacks, because the gradient magnitude becomes very small. 

The above formulation does not specify how to generate $x', y'$. Mathematically, the optimization problem can be written as
\begin{equation}
\max_{x' \in \mathcal{S}_x, y' \in \mathcal{S}_y} \ L(x', y';\theta).
\end{equation}
Here $\mathcal{S}_x$ is defined by~(\ref{s_x}), and  $\mathcal{S}_y$ is defined as
\begin{equation}
\mathcal{S}_y = \{z\,|\, z\in B(y, \epsilon_y), \ z  \geq 0, \ {\textstyle\sum}_i{z_i} = 1\}.
\end{equation}
The final formulation for adversarial training is as follows
\begin{equation}\label{eq:formulation}
\min_{\theta}\big[\max_{x' \in \mathcal{S}_x, y' \in \mathcal{S}_y} L(x', y';\theta)\big],
\end{equation}
where $(x,y) \!\!\!\sim\!\!\! \mathcal{D}$.
Our simple strategy to solve (\ref{eq:formulation}) is to decompose it into two separate sub-problems, and optimize over $x'$ or $y'$  conditioned on the other respectively. After obtaining $x',y'$, we use them in place of the original $x,y$ as the training data points and optimize over $\theta$. In other words, the training batch only contains adversarially perturbed images. In the next two subsections, we will describe the solution to each sub-problem respectively.

\subsection{Generating Adversarial Labels}
We first study how to compute the adversarial label $y'$ given the original data point $x,y$. We need to solve
\begin{equation}
\max_{y' \in \mathcal{S}_y} \ L(x, y';\theta) \label{optimize_y}.
\end{equation}
Here the original groundtruth $y$ is a one-hot vector, i.e., \mbox{$y_c\!=\!1$} and $y_k\!=\!0,k\!\neq \!c$. We use $k$ to denote the class index and $c$ to denote the groundtruth class.
The most straightforward idea is to use the one-step PGD
\begin{align}
y' & = \Pi_{\mathcal{S}_y}\big(y +  \alpha \nabla_yL(x,y;\theta)\big), \\ 
  & \nabla_{y_k}L(x,y;\theta) = -\log(p_k).
\end{align}
Here the $\Pi_{\mathcal{S}_y}(\cdot)$ operator projects the input into the feasible region $\mathcal{S}_y$. Basically it ensures that the adversarial label $y'$ is 1) in $B(y, \epsilon_y)$  and 2) in the probability simplex. Next we propose a heuristic solution to achieve both. 
In the following, we will use some short notations.
$$v_k = \nabla_{y_k}L(x,y;\theta), \ v_{MC} = \min_{k \neq c} v_k, \ v_{LL} = \max_{k \neq c} v_k.$$
Here ``MC" (most confusing) corresponds to the non-grountruth class with the highest probability, and ``LL" (least likely) corresponds to the non-grountruth class with the lowest probability. 
The idea is that we decrease $y_c$ from 1 to $1 - \epsilon_y$, and distribute $\epsilon_y$ to other non-groundtruth classes. The share for each class is based on their respective gradient $\nabla_{y_k}L(x,y;\theta)$, while the share for the MC class (i.e., $y'_{MC}$) is set to be very small. This way, we can obtain
\begin{equation}
y'_k = \frac{\epsilon_y}{n - 1} \cdot \frac{v_k - v_{MC} + \gamma}{\frac{\sum _{k \neq c} v_k}{n - 1} - v_{MC} + \gamma}, \quad k \neq c.
\label{solution}
\end{equation}
Here $\gamma$ is a very small value, e.g., $0.01$. Please refer to the supplementary material for another heuristic solution and the comparison. It is easy to see that if the gradient of non-groundtruth classes are equal, the second multiplicative term becomes 1 and we then obtain
\begin{equation}
y'_k = \frac{\epsilon_y}{n - 1}, \quad k \neq c.
\end{equation}
This is exactly the label smoothing~\cite{szegedy2016rethinking}. In other words, label smoothing can be thought of as an adversarial perturbation of the groundtruth label.

Note that $\epsilon_y$ controls the perturbation budget of $y$. We are interested in finding the largest $\epsilon_y$ that leads to the most adversarially perturbed label. The idea is that we want to keep the probability of the groundtruth class (i.e., $y'_c$) at least $\beta$ times larger than the maximal probability over non-groundtruth classes (i.e., $y'_{LL}$). Mathematically, we want
\begin{equation}
y'_c \geq \beta \cdot \max_{k \neq c}{y_k'}.
\end{equation}
Solving for the following equation
\begin{equation}
1 - \epsilon_y \geq \frac{\beta \epsilon_y}{n - 1} \cdot \frac{v_{LL} - v_{MC} + \gamma}{\frac{\sum _{k \neq c} v_k}{n - 1} - v_{MC} + \gamma},
\end{equation}
we obtain
\begin{equation}
\epsilon_y \leq \frac{1}{1 + \frac{\beta}{n - 1}\frac{v_{LL} - v_{MC} + \gamma}{\frac{\sum _{k \neq c} v_k}{n - 1} - v_{MC} + \gamma}}. 
\label{epsilon_y}
\end{equation}

Next we consider two extreme cases. 

1) The probabilities of non-groundtruth classes are evenly distributed, i.e., label smoothing. In this case, $v_{LL} = v_{MC}$. Then we have 
\begin{equation}
\epsilon_y = \frac{1}{1 + \frac{\beta}{n - 1}}.
\end{equation}
Take CIFAR10 for example ($n\!\!= \!\!10$). We have \mbox{$\epsilon_y \!=\! 0.1$}, \mbox{$\beta\!=\!81$}, or $\epsilon_y\!=\!0.5$, $\beta\!=\!9$, or $\epsilon_y\!=\!0.9$, $\beta\!=\!1$.

2) The probabilities of non-groundtruth classes are centered on one class. In this case, $v_{LL} \!=\! {\textstyle\sum}_{k \neq c} v_k$, $v_{MC}\!=\!0$. Then we have
\begin{equation}
\epsilon_y = \frac{1}{1 + \beta \frac{v_{LL} + \gamma}{v_{LL} + (n - 1) \gamma}} \approx \frac{1}{1 + \beta}.
\end{equation}
Note that $\gamma$ is usually very small, e.g. $\gamma \!=\! 0.01$.

We can see that given the multiplier $\beta$, the range of perturbation budget $\epsilon_y$ is
\begin{equation}
\epsilon_y \in (\frac{1}{1 + \beta}, \frac{1}{1 + \frac{\beta}{n - 1}}].
\end{equation}
Note that we only need to specify a proper $\beta$. As a special case, $\beta=\infty$ corresponds to the original one-hot label.

\subsection{Generating Adversarial Images}
Next we study how to compute the adversarial image $x'$ given the original data point $x,y$. Mathematically, we need to solve the problem
\begin{equation}
\max_{x' \in \mathcal{S}_x} \ L(x', y;\theta).
\label{adv_x}
\end{equation}
This is the adversarial attack problem. For non-targeted attacks, we directly maximize (\ref{adv_x}). The downside of using non-targeted attacks is label leaking~\cite{kurakin2016scale}. This is because during training the model implicitly learns to infer the true label from the adversarial perturbation (gradient). In other words, the model smartly finds a shortcut (degenerate minimum~\cite{tramer2017ensemble}) towards the local optima. A more general and severe problem is gradient masking~\cite{papernot2016distillation}. It refers to the fact that the loss surface of the model is very jagged, and so it becomes harder for the attackers to find good gradient during the iterative attack. As demonstrated in~\cite{athalye2018obfuscated}, gradient masking (a.k.a. gradient obfuscation) gives a false sense of robustness, and the model gets broken in the face of strong attacks with large number of iterations. 

Recently, two techniques were proposed to reduce or avoid gradient masking problem: 1) using multi-step PGD~\cite{madry2017towards}; 2) using an ensemble of models to generate adversarial examples~\cite{tramer2017ensemble}. However, the effectiveness comes with expensive time cost~\cite{madry2017towards} or memory cost~\cite{tramer2017ensemble}. Since one of our design consideration is speed, in this paper, we focus on two simple techniques: 1) using targeted attack~\cite{kurakin2016scale}; 2) adding random noise as in Eq.~(\ref{random})~\cite{tramer2017ensemble,madry2017towards}.


As for targeted attack, in~\cite{kurakin2016scale}, the authors used the Least Likely (LL) class as the targeted class, i.e., 
\begin{equation}
y' = \text{argmax}_{\hat{y} \neq y} L(x, \hat{y}; \theta).
\end{equation}
Differently, here in this paper, we use the Most Confusing (MC) class as the targeted class, i.e., 
\begin{equation}
y' = \text{argmin}_{\hat{y} \neq y} L(x, \hat{y}; \theta).
\end{equation}
In order for fast training, we use one-step PGD (in the experiment on the difficult ImageNet we use two-step). Note that the update equations (\ref{random}) and (\ref{PGD}) are for non-targeted attacks. For targeted attacks, we simply replace the groundtruth label $y$ by the targeted label $y'$, and also replace the plus sign by the minus sign, in order to minimize the loss with respect to the targeted label.



\section{Experiments}
{\flushleft\textbf{Dataset and Network.}} In the experiment, we use CIFAR10~\cite{krizhevsky2009learning}, and SVHN~\cite{netzer2011reading} and the large-scale ImageNet~\cite{deng2009imagenet}. We do not use MNIST~\cite{lecun1998gradient} because it is not a good testbed due to the near-binary nature of the images~\cite{tramer2017ensemble}. For CIFAR10 and SVHN, we use Wide ResNet~\cite{zagoruyko2016wide} (WRN-28-10). For ImageNet, we use ResNet family~\cite{he2016identity}. Most of the diagnostic experiments are conducted on CIFAR10, because it is currently the most commonly used dataset for adversarial training.

{\flushleft\textbf{Evaluation.}} Based on the amount of knowledge that attackers have, there are several types of attacks:
i)~Gradient-based (white-box), where the attackers have full knowledge of the model (structure, parameters);
ii)~Score(logit)-based, where the attackers know the score/logit vector of the model (e.g. SPSA~\cite{uesato2018adversarial}, ZOO~\cite{chen2017zoo}); 
iii)~Decision-based, where the attackers only know the predicted class (e.g. boundary attack~\cite{brendel2017decision}).
Note that the more information the attackers have, the stronger the adversarial attacks will be. In the experiment we use the strongest gradient-based white-box attacks.

For CIFAR10 and SVHN, we follow the evaluation setup in~\cite{madry2017towards}. Specifically, the test perturbation budget is 8 pixels. In order to use strong attacks to evaluate the model robustness, we 1) always use non-targeted attack, and 2) drop random start for one-step PGD attack (i.e. FGSM), and use random start for multi-step PGD attack. Also, for one-step attack, the step size is the perturbation budget (i.e., 8 pixels), and for multi-step attack, the step size is 2 pixels.

For ImageNet, we follow the evaluation setup in~\cite{kannan2018adversarial,xie2018feature}. Specifically, the test perturbation budget is 16 pixels. We use the non-targeted attack, and the targeted attack where the label is randomly selected. The step size is 1 pixel, expect for PGD10 attack where the step size is 1.6 pixel. 

{\flushleft\textbf{Implementation Details.}} For CIFAR10 and SVHN, we largely follow the released code in~\cite{madry2017towards}. The learning rate schedule is [0.1, 0.01, 0.001] for CIFAR and [0.01, 0.001, 0.0001] for SVHN. For the short training, the decay epoch schedule is [60, 90, 100]. And for the long training, the epoch schedule [100, 150, 200]. In all the tables, the models of long training are postfixed by ``+".
For ImageNet, we use the Tensorpack package and perform distributed adversarial training with 8 GPUs. We largely follow the code for distributed training ResNet models on ImageNet. The learning rate schedule is [0.1, 0.01, 0.001, 0.0001], and the decay epoch schedule is [30, 60, 90, 100]. For ResNet50, the training takes about 2 days on a DGX machine.
We plan to release the code and models after this work is published.

\begin{table}[t!]
\centering
\tabcolsep=0.05cm
\begin{tabular}{ |c||c|c|c|c|c|c| }
\hline
\multirow{2}{*}{Acc.(\%)} & \multirow{2}{*}{clean}
                  & \multicolumn{2}{c|}{FGSM}
                  & \multicolumn{2}{c|}{CE7}
                  & \multirow{2}{*}{black} \\
\cline{3-6}
 &  & \multicolumn{1}{c|}{w.o. RS} 
                 & \multicolumn{1}{c|}{RS}
                 & \multicolumn{1}{c|}{w.o. RS}
                 & \multicolumn{1}{c|}{RS}
                 &  \\
\hline 
FGSM   & 55.2 & 99.1 & 68.6 & 0.0 & 0.0 & 56.2 \\
\hline
R-FGSM & 89.8 & 55.8 & 63.6 & 46.4 & 48.0 & 88.0 \\
\hline
LL     & 92.6 & 97.9 & 86.2 & 0.0  & 0.0 & 80.9 \\
\hline
R-LL   & 91.4 & 46.6 & 56.6 & 34.1 & 36.0 & 88.2 \\
\hline
MC     & 86.4 & 70.7 & 73.0 & 37.6 & 40.3 & 84.3 \\
\hline
R-MC   & 89.9 & 62.6 & 70.2 & 46.8 & 48.4 & 87.1 \\
\hline
\end{tabular}
\caption{\small The classification accuracy of one-step adversarially trained models, using different attacks, and, with or without random start. The models trained using random start are prefixed with ``R-".  We see that random start and MC targeted attack are effective at preventing label leaking and gradient masking problem.}
\label{random_start}
\end{table}


\begin{table}[t!]
\centering
\tabcolsep=0.06cm
\begin{tabular}{ |c||c|c|c|c|c|c| }
\hline
\multirow{2}{*}{Acc.(\%)}& \multicolumn{2}{c|}{FGSM}
                  & \multicolumn{2}{c|}{MC}
                  & \multicolumn{2}{c|}{LL} \\
\cline{2-7}
      & \multicolumn{1}{c|}{w.o. RS} 
                 & \multicolumn{1}{c|}{RS}
                 & \multicolumn{1}{c|}{w.o. RS}
                 & \multicolumn{1}{c|}{RS}
                 & \multicolumn{1}{c|}{w.o. RS}
                 & \multicolumn{1}{c|}{RS} \\
\hline
R-FGSM  & 55.8 & 63.6 & 55.4 & 63.6 & 75.5 & 79.8  \\
\hline
R-LL    & 46.6 & 56.6 & 44.0 & 55.6 & 70.7 & 76.4  \\
\hline
R-MC    & 62.6 & 70.2 & 63.9 & 71.3 & 80.1 & 83.8  \\
\hline
\end{tabular}
\caption{\small The classification accuracy of three attacks, i.e., FGSM attack, LL targeted attack and MC targeted attack, with or without random start. The rows correspond to different adversarially trained models. We see that MC targeted attack has similar strength as FGSM attack, and both are much stronger than LL targeted attack.}
\label{attack_strength}
\vspace{-0.3cm}
\end{table}


\subsection{Random Start and MC Targeted Attack}
In this subsection, we demonstrate that for one-step adversarial training, random start and MC targeted attack are effective at preventing label leaking and gradient masking problem. This diagnostic experiment is conducted on CIFAR10. To this end, we evaluate three different ways of generating adversarial examples during training: FGSM attack, LL targeted attack, and MC targeted attack. For each option, we perform one-step adversarial training with or without random start. This leads to six adversarially trained models in total. The perturbation budget is 8 pixels in training. The results are shown in Table~\ref{random_start}. The rows correspond to different models, where the prefix ``R" means that random start is used. The columns correspond to non-targeted attacks using one-step (FGSM) or 7-step (CE7, CE is short for Cross-Entropy), with or without random start (denoted by ``RS"). The last column corresponds to the black-box attack using the undefended model and FGSM attack (w.o. RS). Firstly we see that the target models trained by FGSM and LL suffer badly from the label leaking problem because the accuracy against FGSM attack is even higher than the clean accuracy. But this is just false robustness and the accuracy drops to zero under CE7 attacks. Next, after applying random start, R-FGSM and R-LL become quite robust, demonstrating random start helps the model avoid the label leaking problem during adversarial training. Lastly, we see that the model R-MC performs the best, while R-LL performs the worst, against FGSM and CE7 attacks.

We hypothesize that the adversarially trained mode by one-step LL targeted attack is weak, because the LL targeted attack is weak by itself. Table~\ref{attack_strength} shows the strength of these three attacks using one-step PGD. The rows correspond to different models trained using random start. The columns correspond to three attacks with or without random start. We see that MC targeted attack is roughly as strong as FGSM attack, and both of them are much stronger than LL targeted attack. This is probably because it is usually hard to slightly manipulate the original image so that it becomes a visually very different class. Therefore, we argue for using MC targeted attack during adversarial training because 1) it is much stronger that LL targeted attack; 2) FGSM attack risks label leaking and gradient masking problem.

We briefly summarize the role of random start. In training, it effectively prevents the label leaking and gradient masking problem, but in attack, it weakens the strength for one-step PGD attack (shown in Table~\ref{attack_strength}). As we will show later, random start has very little effect for multi-step PGD attack, especially when the number of steps becomes large. 

\begin{table}[t!]
\centering
\tabcolsep=0.06cm
\begin{tabular}{ |c||c|c|c|c| }
\hline
Acc.(\%) & clean & FGSM & CE7 & CE20 \\
\hline
R-FGSM           & 89.8 & 55.8 & 48.0 & 42.9  \\
\hline
R-FGSM-LS ($\epsilon_y=0.5$) & 89.1 & 62.0 & 54.6 & 49.0  \\
\hline
R-MC             & 89.9	& 62.6 & 48.4 & 43.4  \\
\hline
R-MC-LS ($\epsilon_y=0.5$)   & 91.1	& 70.6 & 59.2 & 53.3  \\
\hline
R-MC-LS+ ($\epsilon_y=0.5$)   & {\bf 91.8} & {\bf 71.4} & 62.7 & 55.9  \\
\hline
R-MC-LA ($\beta=9$)     & 90.7	& 69.6 & 59.9 & 55.3  \\
\hline
R-MC-LA+ ($\beta=9$)     & 91.2 & 70.7 & {\bf 63.0} & {\bf 57.8}  \\
\hline
\hline
Madry~\cite{madry2017towards}  & 87.3 & 56.1 & 50.0 & 45.8  \\
\hline
Madry*     & 88.0 & 57.0 & 51.2 & 47.6  \\
\hline
Madry-LA   & 86.8 & 63.4 & 57.8 & 53.2  \\
\hline
Madry-LA+  & 87.5 & 65.9 & 61.3 & 57.5  \\
\hline
\end{tabular}
\caption{\small The classification accuracy of R-MC-LA models and variants under various white-box attacks on CIFAR10.}
\label{cifar10}
\vspace{-0.3cm}
\end{table}

\subsection{CIFAR10 Dataset}
In this subsection, we report results against white-box attacks on CIFAR10~\cite{krizhevsky2009learning} dataset. It has 10 classes, 50K training images (5K per class) and 10K test images. As summarized above, to generate adversarial images, we use random start and MC targeted attack (the perturbation budget is 8 pixels). To generate adversarial labels, we use (\ref{epsilon_y}) to compute the budget $\epsilon_y$ and (\ref{solution}) to compute the adversarially perturbed labels $y'$. The resulting model is denoted as R-MC-LA where LA stands for label adversary. We also experiment with label smoothing (LS for short), a special case of adversarial label, and denote this model by R-MC-LS. Our baseline is the mult-step adversarial training method by~\cite{madry2017towards}. We report the original numbers in their paper, denoted by Madry, and also report the numbers by our reproduced model, denoted by Madry*. The accuracy against various steps of PGD attacks is given in Table~\ref{cifar10}. We see that perturbing labels during training significantly improves the robustness over the baseline. Label smoothing (R-MC-LS) works pretty well, and label adversary (R-MC-LA) is even better. Besides, we also apply label adversary to the multi-step adversarial training. We see that the resulting models, denoted by Madry-LA and Madry-LA+, significantly improves the original version, further verifying the effectiveness of label adversary. Interestingly, R-MC-LA(+) performs even better than Madry-LA(+). Lastly, we observe that longer training is helpful to all models.

\begin{figure}[t!]
\vspace{-0.65cm}
\centering
\includegraphics[width=0.45\textwidth]{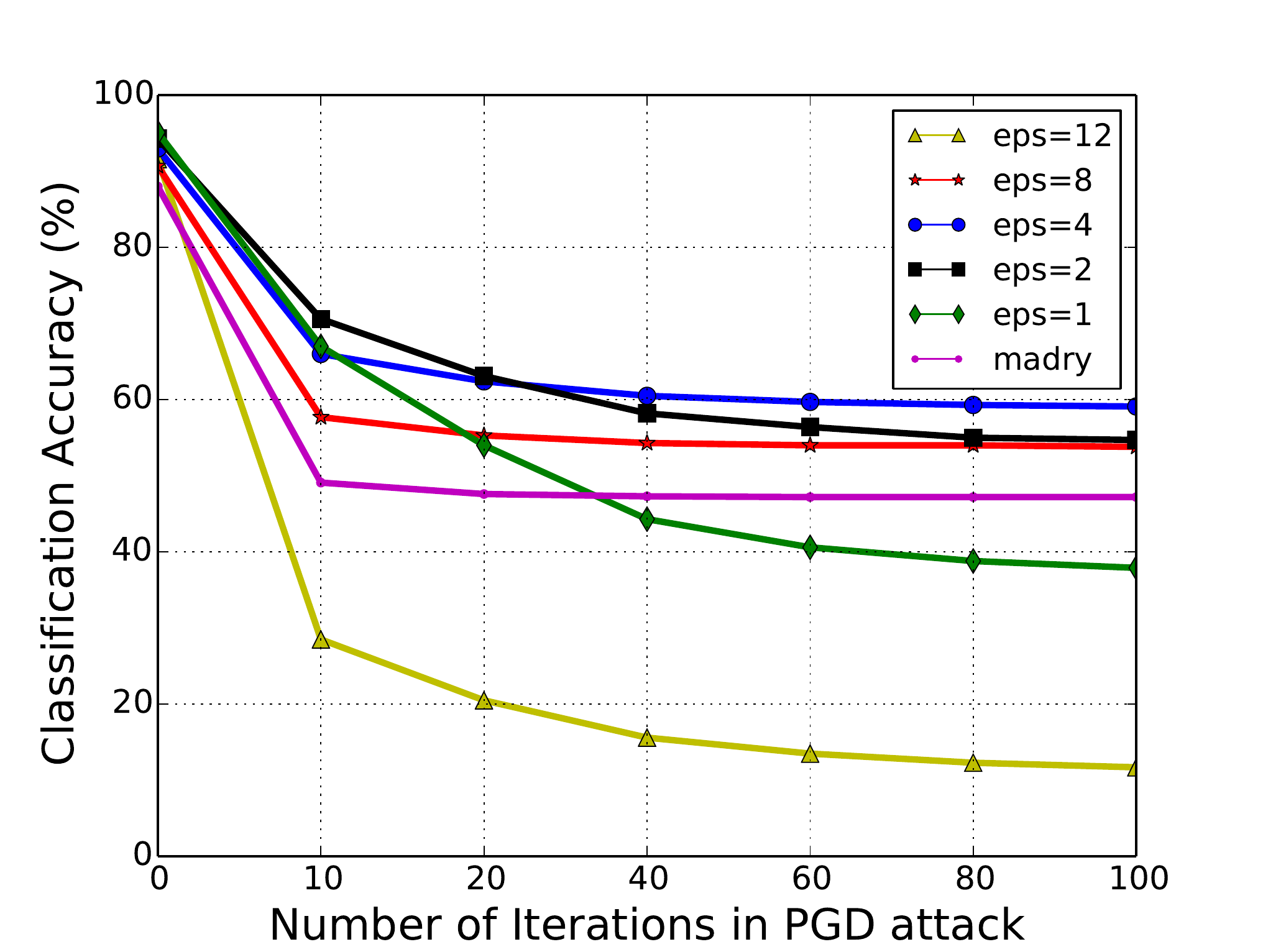}
\caption{\small The classification accuracy of the proposed R-MC-LA models against white-box PGD attacks of different number of iterations on CIFAR10. The models are trained by using different perturbation budget. We use $\beta=9$.}
\label{num_steps}
\end{figure}

\begin{table}[t!]
\centering
\tabcolsep=0.06cm
\begin{tabular}{ |c||c|c|c|c|c|c| }
\hline
Acc.(\%) & \small clean & \footnotesize CE20 & \footnotesize CW20 & \footnotesize CE100 & \footnotesize CW100 & \footnotesize CW200   \\
\hline
\footnotesize R-MC-LA ($\epsilon_x=8$)     & 90.8 & 54.6 & 53.7 & 52.9 & 51.9 & 51.7 \\
\hline
\footnotesize R-MC-LA+ ($\epsilon_x=8$)    & 91.0 & 57.5 & 56.2 & 55.2 & 53.8 & 53.6 \\
\hline
\footnotesize R-MC-LA ($\epsilon_x=4$)     & {\bf 93.0} & 63.1 & 61.5 & 60.1 & 58.0 & 57.6 \\
\hline
\footnotesize R-MC-LA+ ($\epsilon_x=4$)    & 92.9 & {\bf 66.9} & {\bf 64.2} & {\bf 63.7} & {\bf 60.7} & {\bf 60.3} \\
\hline
\hline
Madry*  & 88.0 & 47.6 & 48.6 & 47.2 & 48.1 & 48.1 \\
\hline
\end{tabular}
\caption{\small The classification accuracy of the proposed R-MC-LA models under various white-box attacks on CIFAR10. To rule out randomness, the numbers are averaged over 3 independently trained models. We use $\beta=9$.}
\label{budget}
\vspace{-0.3cm}
\end{table}

\subsubsection{Different Perturbation Budgets during Training}
Next we study whether using larger perturbation budget during training leads to more robust models. We train models using the budget $\epsilon_x \!\in\! \{1, 2, 4, 8, 12\}$ pixels during training, and we use 8 pixels for evaluation. Figure~\ref{num_steps} shows the classification accuracy with respect to the number of steps in PGD attack. Firstly we observe the general trend that as the number of steps increases, the accuracy drops quickly and then plateaus. Secondly, we find that big budget (i.e., $\epsilon_x\!\!=\!\!12$) or small budget (i.e., $\epsilon_x\!\!=\!\!1$) lead to less robust models. Interestingly, we see that the model trained using $\epsilon_x\!=\!4$ achieves the best robustness. The exact numbers are given in Table~\ref{budget}. Note that to rule out randomness, the numbers are averaged over 3 independently trained models. We also test the attacks using the margin-based CW loss~\cite{carlini2016towards}. For example, CW200 attack means using CW loss and 200 steps PGD. We notice that 1) the baseline model~\cite{madry2017towards} performs similarly against either (cross-entropy-based) CE attack or (margin-based) CW attack; 2) CW attack is more effective than CE attack when attacking our models (R-MC-LA). Furthermore, we evaluate our best model, R-MC-LA+ ($\epsilon_x\!=\!4$), against 1000-step PGD attacks using CE loss and CW loss. The accuracy is {\bf 61.4\%} for CE and {\bf 59.3\%} for CW, which is very close to that against 200-step PGD attack.

\subsubsection{Black-box Attack Evaluation}
We next evaluate our R-MC-LA models against black-box attacks. We use two models to generate the adversarial examples: the undefended model and another randomly initialized R-MC-LA model. All the models are trained using $\epsilon_x=8$. The results are shown in Table~\ref{black}. We see that attacks generated by R-MC-LA model are stronger than those by undefended model, because two independently trained R-MC-LA models share inherent structure. Besides, we see that all the black-box attacks are weaker than white-box attacks (by comparing the accuracy), demonstrating that our models do not suffer from the gradient masking problem.
\begin{table}[t!]
\centering
\tabcolsep=0.06cm
\begin{tabular}{ |c||c|c|c|c|c| }
\hline
\multirow{2}{*}{Acc.(\%)} & \multirow{2}{*}{clean}
                  & \multicolumn{2}{c|}{Undefended}
                  & \multicolumn{2}{c|}{\small another R-MC-LA} \\
\cline{3-6}
 &  & \multicolumn{1}{c|}{FGSM}
                 & \multicolumn{1}{c|}{CE20}
                 & \multicolumn{1}{c|}{FGSM}
                 & \multicolumn{1}{c|}{CE20} \\
\hline
\small R-MC-LA   & 90.7 & 87.8 & 88.8 & 74.4 & 71.0 \\
\hline
\small R-MC-LA+  & 91.2 & 88.5 & 89.9 & 74.6 & 74.4 \\
\hline
\end{tabular}
\caption{The classification accuracy of R-MC-LA models against black-box attacks on CIFAR10. We use $\beta=9$.}
\label{black}
\end{table}

\begin{table}[t!]
\centering
\tabcolsep=0.06cm
\begin{tabular}{ |c||c|c|c|c|c| }
\hline
Acc.(\%) & clean & \small FGSM & \small CE20 & \small CE100 & \small CW100 \\
\hline
\footnotesize R-MC-LA ($\epsilon_x=8$)    & 94.1 & 66.9 & 46.7 & 42.0 & 40.9 \\
\hline
\footnotesize R-MC-LA+ ($\epsilon_x=8$)   & 94.1 & 69.8 & 53.9 & 50.3 & 48.9 \\
\hline
\footnotesize R-MC-LA ($\epsilon_x=4$)    & {\bf 95.7} & 72.6 & 54.4 & 47.2 & 45.7 \\
\hline
\footnotesize R-MC-LA+ ($\epsilon_x=4$)   & 95.5 & {\bf 74.2} & {\bf 63.0} & {\bf 59.1} & {\bf 58.5} \\
\hline
\hline
Madry*   & 91.8 & 61.0 & 43.2 & 42.1 & 43.4 \\
\hline
\end{tabular}
\caption{The classification accuracy of R-MC-LA models under various white-box attacks on SVHN. The models are trained using different perturbation budget. We use $\beta\!=\!9$.}
\label{svhn}
\vspace{-0.3cm}
\end{table}

\subsection{SVHN Dataset}
The SVHN~\cite{netzer2011reading} is a 10-way house number classification dataset. It contains 73257 training images, and 26032 test images. We don't use the additional training images. The results against white-box attacks are shown in Table~\ref{svhn}. Similar to CIFAR10, we see that our models significantly outperform the state-of-the-art results on clean images and against PGD attacks of various strength.

\begin{table}[t!]
\centering
\tabcolsep=0.06cm
\begin{tabular}{ |c||c|c|c|c|c| }
\hline
Acc.(\%) & clean & \small CE10-nt & \small CE100-nt & \small CE10-rd & \small CE100-rd  \\
\hline
\footnotesize R-MC-LA-R50    & 58.9 & 14.9 & 4.0 & 45.8 & 24.5 \\
\hline
\footnotesize R-MC-LA-R101   & 61.9 & 18.0 & 6.3 & 45.8 & 26.0 \\
\hline
\footnotesize R-MC-LA-R152   & 63.9 & {\bf 19.8} & {\bf 7.4} & 46.5 & 26.6 \\

\hline
\hline
\small \cite{kannan2018adversarial}-IncepV3   & {\bf 72.0} & NA & NA & 27.9 & NA \\
\hline
\small \cite{xie2018feature}-R152  & 62.3 & 17.1 & 7.3 & {\bf 52.5} & {\bf 41.7} \\
\hline
\end{tabular}
\caption{The classification accuracy of R-MC-LA models under various white-box attacks on ImageNet. We use $\beta\!=\!100$. The budget is 16 pixels in training and evaluation.}
\label{imagenet}
\vspace{-0.35cm}
\end{table}

\subsection{ImageNet Dataset}
The ImageNet dataset has ∼1.28 million training images with 1000 classes. We use the validation set with 50K images for evaluation. To the best of our knowledge, up to now, there are only two papers that have applied multi-step adversarial training on ImageNet, because it is very computationally expensive. Specifically, the prior art~\cite{kannan2018adversarial} used 53 P100 GPUs and the recent paper~\cite{xie2018feature} used 128 V100 GPUs. We train our models on a DGX machine with only 8 GPUs and it takes us about 2 days. For fair comparison, we use 16 pixels as test perturbation budget. In our experiment, we find that, using one-step attack during training (in this case step size is 16 pixels) suffers severely from label leaking and gradient masking. We also observe the similar problem on CIFAR10 and SVHN, when training with large budgets, e.g., 12 or 16. In order to make our method work for the 16-pixel evaluation setup, we use two-step MC targeted attack (in this case the step size become 8 pixels). In the experiment, we find that training without label adversary performs very bad, further demonstrating the effectiveness of label adversary. In evaluation, we use both the non-targeted attack and the targeted attack where the target label is uniformly randomly selected. Note that the non-targeted attack is much stronger than the randomly targeted attack, so we believe using both will lead to a more reliable robustness evaluation. Table~\ref{imagenet} shows the top-1 accuracy of our method and two baseline methods, where the non-targeted attack is denoted by ``nt", and the randomly targeted attack is denoted by ``rd". We can see that our methods significantly outperform the prior art~\cite{kannan2018adversarial} against CE10-rd attack. Compared with the recent work~\cite{xie2018feature}, our models are better on clean accuracy and against non-targeted attacks, but are worse against randomly targeted attacks. We hypothesize that this may be because that the models in~\cite{xie2018feature} are trained using randomly targeted attack (same as test), and, they are using an-order-of-magnitude more computational resources (30-step PGD during adversarial training). 

\section{Conclusion}
In this paper, we proposed to use the adversarial image and the adversarial label during adversarial training. The adversarial image was generated by one-step or two-step MC targeted attack. The adversarial label was computed by an close-form formula. Comprehensive experiments on CIFAR10, SVHN and ImageNet, against various white-box attacks, demonstrate the effectiveness of our method.

\clearpage

{\small
\bibliographystyle{ieee}
\bibliography{egbib}
}

\clearpage

\appendix
{\huge Appendix}

\section{Introduction}
First, we provide two more experiments on CIFAR10. Then, we provide the results on CIFAR100 dataset. Next, to validate our motivation, we compare the gradient magnitude of different models. At last, we provide another heuristic solution to the problem of generating adversarial labels.

\section{Two More Experiments on CIFAR10}
\begin{table}[h!]
\centering
\tabcolsep=0.06cm
\begin{tabular}{ |c||c|c|c|c|c|c| }
\hline
Acc.(\%) & clean & \small FGSM & \small CE1000 & \small MI & \small Ori-CW & \small DF-l2 \\
\hline
\small R-MC-LA+ &  92.8 & 75.6 & 61.4 & 65.5 & 65.4 / 88.2 & 77.8  \\
\hline
\small TRADES &  84.9 & 61.1 & 56.4 & 58.0 & 81.2 & 81.6  \\
\hline
\end{tabular}
\caption{The classification accuracy against various white-box attacks on CIFAR10. We use training budget $\epsilon_x=4$.}
\label{other}
\end{table}

\subsection{Against Other White-box Attacks}
Table~\ref{other} shows the accuracy against other white-box attacks. We compare with the winner, TRADES~\cite{zhang2019theoretically}, in NeurIPS18 Adversarial Vision Challenge. We use the default settings in the Cleverhands package to generate the attacks. ``MI" refers to the MI-FGSM method~\cite{dong2017boosting}. ``Ori-CW" refers to the original CW attack~\cite{carlini2016towards}, and the two numbers refer to two common sets of hyper-parameters: \{const=100, confid=0, lr=1e-1, max iter=1e2\} / \{const=100, confid=0, lr=1e-2, max iter=1e3\}. ``DF-l2" refers to the DeepFool attack with $l_2$-norm~\cite{moosavi2015deepfool}. We see that our models generally outperform the baseline, except against the DeepFool attack. We note that our network is smaller, and our training method is an-order-of-magnitude faster\footnote{~\cite{zhang2019theoretically} uses WRN-34, and 20-step PGD attack during training.}.

\subsection{Effect of Number of Training Images}
We vary the number of training images per class. The results are shown in Table~\ref{data_size}. This is aligned with the claim in~\cite{schmidt2018adversarially} that adversarial training requires more data than regular training.

\begin{table}[h!]
\centering
\tabcolsep=0.06cm
\begin{tabular}{ |c||c|c|c|c|c| }
\hline
Acc.(\%) & clean & \small FGSM & \small CE20 & \small CE100 & \small CW100 \\
\hline
\small R-MC-LA (5K)     & 90.7  & 69.6  & 55.3  & 53.8  & 52.8 \\
\hline
\small R-MC-LA (2K)     & 85.6  & 56.1  & 42.8  & 41.1  & 40.2 \\
\hline
\small R-MC-LA (0.5K)   & 73.3  & 33.7  & 25.1  & 24.5  & 24.0 \\
\hline
\end{tabular}
\caption{The classification accuracy of R-MC-LA models, trained using different data size on CIFAR10. The number in the parenthesis means the number of images per class. We use $\epsilon_x=8, \beta=9$.}
\label{data_size}
\end{table}

\begin{table}[t!]
\centering
\tabcolsep=0.06cm
\begin{tabular}{ |c||c|c|c|c|c| }
\hline
Acc.(\%) & clean & \small FGSM & \small CE20 & \small CE100 & \small CW100  \\
\hline
\footnotesize R-MC-LA ($\epsilon_x=8$)    & {\bf 68.7}	& 30.5 & 23.2 & 22.7 & 20.6 \\
\hline
\footnotesize R-MC-LA+ ($\epsilon_x=8$)   & 66.2	& 31.3 & 23.1 & 22.4 & 20.0 \\
\hline
\footnotesize R-MC-LA9 ($\epsilon_x=8$)    & {\bf 68.7}	& 33.7 & 23.1 & 22.0 & 20.1 \\
\hline
\footnotesize R-MC-LA9+ ($\epsilon_x=8$)   & 68.2	& {\bf 36.9} & {\bf 26.7} & {\bf 25.3} & 22.1 \\
\hline
\hline
Madry*   & 61.9 & 28.8 & 23.7 & 23.4 & {\bf 24.5} \\
\hline
\end{tabular}
\caption{The classification accuracy of R-MC-LA models against white-box attacks on CIFAR100. The models are trained using different perturbation budget. We use $\beta=11$.}
\label{cifar100}
\end{table}

\section{CIFAR100 Dataset}
In this section we report the results against white-box attacks on CIFAR100~\cite{krizhevsky2009learning} dataset. It has 100 classes, 50K training images and 10K test images. In addition to the basic R-MC-LA models, we also try a slightly modified version, denoted by R-MC-LA9. Specifically, when generating the adversarial label, we distribute the $\epsilon_y$ to the top-9 non-groundtruth classes with largest loss, instead of to all the non-groundtruth classes. This modification brings several percentage gain. The results are shown in Table~\ref{cifar100}. We see that our models outperform the state-of-the-art on clean image and against FGSM, and perform comparably on multi-step attacks. We hypothesize that CIFAR100 is more difficult than CIFAR10 and SVHN for adversarial training because of much fewer images per class.

\begin{table*}[t!]
\centering
\tabcolsep=0.08cm
\begin{tabular}{ |c||c|c|c||c|c|c||c|c|c| }
\hline
\multirow{2}{*}{}  & \multicolumn{3}{c||}{CIFAR10}
                   & \multicolumn{3}{c||}{SVHN}
                   & \multicolumn{3}{c|}{ImageNet} \\
\cline{2-10}
                 & \multicolumn{1}{c|}{undefended} 
                 & \multicolumn{1}{c|}{madry}
                 & \multicolumn{1}{c||}{ours}
                 & \multicolumn{1}{c|}{undefended}
                 & \multicolumn{1}{c|}{madry}
                 & \multicolumn{1}{c||}{ours}
                 & \multicolumn{1}{c|}{undefended}
                 & \multicolumn{1}{c|}{madry}
                 & \multicolumn{1}{c|}{ours} \\
\hline
max     & 349.9  & 1.8     & 1.6     & 267.9  & 15.3  &  3.4    & 0.48    & 0.004 &  0.084 \\
\hline
mean    & 2.3    & 0.022   & 0.0098  & 0.77   & 0.12  &  0.022  & 0.0044  & 0.000038  & 0.00014 \\
\hline
\end{tabular}
\caption{\small Comparison of gradient magnitude, $\|\nabla_xL(x,y;\theta)\|_2^2$, of undefended model, madry's model, and our model, on three datasets (averaged over all test / validation images). The gradient is taken w.r.t. the original image range [0, 255], instead of [-1, 1], so the numbers are 127.5 times smaller than Table 2 in the paper. For CIFAR10 and SVHN, all models are run {\bf three times} to average out randomness.}
\label{grad_magn}
\end{table*}

\section{Gradient Magnitude}
Table~\ref{grad_magn} provides the gradient magnitude results on three datase. For CIFAR10 and SVHN, madry's method is trained using PGD7-2 with budget 8 pixels. For ImageNet, madry's method is trained using PGD10-3 with budget 16 pixels. From the table, we see that adversarially trained models, including madry's and ours, leads to much smaller gradient magnitude (one or two order-of-magnitude), compared to undefended models. This correlates with our hypothesis that there may be a link between small gradient magnitude and adversarial robustness. Besides, we see that madry's and ours are comparable (particularly on CIFAR10 and SVHN). Note that although gradient magnitude confidently distinguish undefended and adversarially trained models, it is not a precise indicator of robustness between adversrially trained models. Currently, only from the gradient magnitude, we cannot confidently tell which one is more robust. 
So we have to compare and report their accuracy. 
Finding precise indicator for adversarial robustness is an active and unsolved research topic.

\section{Another Solution to Generating Adversarial Labels}
In this section, we provide another heuristic solution to the problem of generating adversarial labels
\begin{equation}
\max_{\|y' - y\|_{\infty} \leq \epsilon_y} \ L(x, y';\theta).
\label{adversarial_y}
\end{equation}
Here the original groundtruth $y$ is a one-hot vector, i.e., $y_c=1$ and $y_k=0,k \neq c$.

In the main paper, the heuristic is to distribute the $\epsilon_y$ to non-groundtruth classes while keeping the share of the MC class very small. Specifically, the share is proportional to the gradient of each class subtracted by the minimal gradient (which corresponds to the MC class). Here, we propose another simpler heuristic, which is that the share is directly proportional to the respective gradient. We can then easily obtain the formula
\begin{equation}
y'_k = \frac{\epsilon_y v_k}{\sum _{k \neq c}{v_k}}, k \neq c.
\label{solution_supp}
\end{equation}
Note that we use $v_k$ to denote $\nabla_{y_k}L(x,y;\theta)$ for short. By using the following condition
\begin{equation}
y_c \geq \beta \max_{k \neq c}{y_{k'}},
\end{equation}
we can solve for the largest budget $\epsilon_y$
\begin{equation}
\epsilon_y \leq \frac{1}{1 + \frac{\beta v_{max}}{\sum _{k \neq c}{v_k}}}. 
\end{equation}
Note that this solution is an exact application of gradient ascent
\begin{equation}
y'_k = y_k + \alpha \nabla_{y_k}L(x,y;\theta), k \neq c, \\ 
\end{equation}
where
\begin{equation}
\alpha = \frac{1}{\sum _{k \neq c}{v_k} + \beta v_{max}}.
\end{equation}

\begin{table}[t!]
\centering
\tabcolsep=0.06cm
\begin{tabular}{ |c||c|c|c|c|c| }
\hline
Acc.(\%) & clean & \small FGSM & \small CE20 & \small CE100 & \small CW100 \\
\hline
\footnotesize R-MC-LA (main)    & 90.8 & 69.3 & 54.6 & 52.9 & 51.9 \\
\hline
\footnotesize R-MC-LA+ (main)   & 91.0 & 70.3 & {\bf 57.5} & {\bf 55.2} & {\bf 53.8} \\
\hline
\footnotesize R-MC-LA (sup)     & 90.2 & 70.9 & 53.2 & 51.1 & 49.9 \\
\hline
\footnotesize R-MC-LA+ (sup)    & {\bf 91.5} & {\bf 71.4} & 57.2 & 54.1 & 51.5 \\
\hline
\end{tabular}
\caption{The classification accuracy of the proposed R-MC-LA models under various white-box attacks on CIFAR10. The source models are trained using two solutions for generating the adversarial labels. We use $\beta=9$ and $\epsilon_x=8$ during training and in evaluation.}
\label{cifar}
\vspace{-0.3cm}
\end{table}

We favor the solution used in the main paper over this solution (\ref{solution_supp}) for two reasons. Firstly, from the optimization point of view, the solution in the main paper leads to a higher (better) objective value for the maximization problem~(\ref{adversarial_y}), because it distributes more shares to the classes with larger gradient. Secondly, the solution in the main paper leads to a smaller $y'_{MC}$ (proof given below). Note that the adversarial image used in training is generated by the MC targeted attack. Using a smaller $y'_{MC}$ will suppress the network to predict large probability on the MC class, thus better focusing on predicting large probability on the groundtruth class. The results achieved by these two solutions are shown in Table~\ref{cifar}, where ``main" refers to using the solution in the main paper, and ``sup" refers to using the solution~(\ref{solution_supp}) in the supplementary material. We can see that ``main" is slightly better than ``sup" against multi-step PGD attacks.

Lastly we provide the proof. From the solution in the main paper, we have
\begin{equation}
\begin{split}
& y'_{MC, main} = \\
& \frac{\gamma}{\sum _{k \neq c} v_k - (n - 1)(v_{MC} - \gamma) + \beta(v_{LL} - v_{MC} + \gamma)}.
\end{split}
\label{one}
\end{equation}
From the solution~(\ref{solution_supp}), we have
\begin{equation}
y'_{MC, sup} = \frac{v_{MC}}{\sum _{k \neq c}{v_k} + \beta v_{LL}}.
\label{two}
\end{equation}
The sufficient and necessary condition of
\begin{equation}
y'_{MC, main} < y'_{MC, sup}
\end{equation}
is
\begin{equation}
(n-1)v_{MC} + \beta v_{MC} < \sum _{k \neq c}{v_k} + \beta v_{LL},
\end{equation}
which is obviously true. This is because the left is smaller than the right on both the first term and the second term respectively.

\end{document}